\documentclass[]{spie} 
 
\usepackage{longtable}
\usepackage[misc]{ifsym}
\usepackage{booktabs}
\usepackage{amsmath,amsfonts,amssymb}
\usepackage{graphicx}
\usepackage[colorlinks=true, allcolors=blue]{hyperref}
\usepackage{multirow}
\usepackage{xcolor}
\title{Towards a Psychological Generalist AI: A Survey of Current Applications of Large Language Models and Future Prospects}

\author[a]{Tianyu He\textsuperscript{*}}
\author[b]{Guanghui Fu\textsuperscript{*}}
\author[a]{Yijing Yu}
\author[a]{Fan Wang}
\author[c]{Jianqiang Li}
\author[c]{Qing Zhao}
\author[c]{Changwei Song}
\author[c]{Hongzhi Qi}
\author[a]{Dan Luo}
\author[a]{Huijing Zou}
\author[a]{\Letter\ Bing Xiang Yang}

\affil[a]{Center for Wise Information Technology of Mental Health Nursing Research, School of
Nursing, Wuhan University, Wuhan, China}
\affil[b]{Sorbonne Université, Institut du Cerveau - Paris Brain Institute - ICM, CNRS, Inria, Inserm, AP-HP, Hôpital de la Pitié Salpêtrière, Paris, France}
\affil[c]{School of Software Engineering, Beijing University of Technology, Beijing, China}

\authorinfo{\textsuperscript{*} These authors contributed to the work equllly and should be regarded as co-first authors. Further author information: (Send correspondence to yangbx@whu.edu.cn)}

\begin{document} 
\maketitle
%Detailed abstract for technical review purposes (200-300 words).

\begin{abstract}
The complexity of psychological principles underscore a significant societal challenge, given the vast social implications of psychological problems. Bridging the gap between understanding these principles and their actual clinical and real-world applications demands rigorous exploration and adept implementation. In recent times, the swift advancement of highly adaptive and reusable artificial intelligence (AI) models has emerged as a promising way to unlock unprecedented capabilities in the realm of psychology. This paper emphasizes the importance of performance validation for these large-scale AI models, emphasizing the need to offer a comprehensive assessment of their verification from diverse perspectives. Moreover, we review the cutting-edge advancements and practical implementations of these expansive models in psychology, highlighting pivotal work spanning areas such as social media analytics, clinical nursing insights, vigilant community monitoring, and the nuanced exploration of psychological theories. Based on our review, we project an acceleration in the progress of psychological fields, driven by these large-scale AI models. These future generalist AI models harbor the potential to substantially curtail labor costs and alleviate social stress. However, this forward momentum will not be without its set of challenges, especially when considering the paradigm changes and upgrades required for medical instrumentation and related applications.
\end{abstract}
% Some general suggestions: 
% 介绍文献或者系统时，可以写成: Fu et al.~\cite{} introduce/proposed/... 
% 引用文献时，请加上波浪线，这样可以避免渲染时候换行问题。From \cite{} to ~\cite{}
% 在最后面加入新章节缩写，来总结全部出现的缩写。注意通常缩写出现第一次时候用全称(缩写)，后文直接用缩写即可。但这要根据情况而定，在不影响阅读的情况下。例如有时突然出现不熟悉的缩写，或者全称出现在很远的地方，会影响阅读效果，所以此时应该用全称。考虑到引文过多，如果有的缩写只出现一次，例如EI: Emotional intelligence，EU: Emotion understanding等，不用写缩写，直接全程就好。缩写的目的是不影响阅读的前提下，简化表述。
% 引用文献支持观点是必要的，但是不必在一个段落里频繁的每句引用，如果中间都是对该文献的描述或观点引用，那么只出现在第一次即可。引用的作用是支撑你的说法，但是过于频繁的引用单个文献是不必要的，因为读者已经知道该观点的来源
% 主要观点有时需要引用支撑，此时尽量引用权威且已同行评审过的期刊文献，尽可能避免Arxiv预印版。但是由于该领域较为新颖，大多数都是Arxiv，所以我们引用Arxiv是可以的。但是记得大观点的支撑尽量来源于权威期刊。
% 全文术语要统一，例如是ChatGPT 还是GPT-3.5。因为各个文献中的叫法不一致，我觉得统一为GPT-3.5或GPT-4就好。根据OpenAI的官方说法，ChatGPT指的是GPT-3.5，而高级版本指的是GPT-4，不带"Chat"

\section{Introduction} \label{sec:intro} 
Mental health conditions are prevalent across the globe. However, many societies and their corresponding health and social systems overlook mental health, failing to offer the requisite care and support. Consequently, countless individuals worldwide endure these challenges silently, face human rights violations, or encounter adversities in their daily lives~\cite{who2022world}.
Psychological well-being and understanding are of paramount importance in today's rapidly evolving world. Across the globe, individuals grapple daily with the dynamic nature of their psychological states, influenced by both external and internal factors. Whether it's individuals expressing their sentiments on social media~\cite{braghieri2022social}, students and teachers confronting mental health challenges in academic settings~\cite{lever2017school}, or the overarching demand for professional psychological counselors~\cite{gruber2021mental}, the implications are evident—mental health matters. Prompt interventions, especially in critical situations such as potential suicides, can save lives~\cite{dueweke2018suicide, bolton2015suicide}. Moreover, timely support and guidance can ameliorate numerous other psychological challenges, ensuring healthier communities~\cite{lattie2020opportunities}.

The realm of artificial intelligence (AI) has witnessed tremendous advancements in recent years, offering promising tools to address these pressing psychological concerns\cite{graham2019artificial, lee2021artificial}. Deep learning techniques, backed by expert annotations, have been heralded for their impressive performance in various applications~\cite{su2020deep, kim2021machine}. Yet, the heavy reliance on extensive annotations introduces significant costs, which often becomes a barrier for many institutions and researchers~\cite{zhang2022natural}.

Enter the era of large language models (LLMs), epitomized by models like GPT~\cite{openai2023chatgpt3, openai2023gpt4}. These models, pre-trained on vast datasets, leverage intricate neural networks and techniques from reinforcement and supervised learning~\cite{zhao2023survey}. Their ability to interface with humans through prompts allows them to address a multitude of tasks, thereby circumventing the need for extensive expert-labeled data~\cite{liu2023pre}. Not only does this foster a spirit of experimentation in psychological research, but the inherent conversational proficiency of these models positions them as empathetic listeners and virtual companions~\cite{ziems2023can}.

While the psychological domain is undoubtedly poised for a paradigm shift due to these AI advancements, the horizon holds even more promise~\cite{demszky2023using, thirunavukarasu2023large}. The emergence of multi-modal generalist AI models—capable of integrating data from various modalities—foreshadows an era where AI can intuitively grasp and execute diverse human tasks\cite{fei2022towards, moor2023foundation}. These models can discern subtle psychological shifts, hypothesize about uncharted psychological territories leveraging multidisciplinary knowledge, and even pioneer novel paradigms under human guidance.

Generalist AI in psychology promises a future where machines can not only model but also predict forthcoming psychological dynamics. While the journey to a holistic generalist AI remains an ongoing endeavor, current strides in the field are illuminating the path~\cite{van2023global}. In this paper, we provide a comprehensive review of the latest developments in the utilization of LLMs for psychological research. Through a systematic exploration of application domains, target populations, and the extraction of psychological constructs, we envision and articulate the transformative potential of generalist AI in the realm of psychology.

The structure of this paper is outlined as follows. Initially, in Section~\ref{sec:generalist}, we introduce the concept of generalist AI, providing a foundational understanding for readers. This is followed by Section~\ref{sec:evaluation}, which delves into the critical aspect of model validation, a fundamental prerequisite for practical application. Next, in Section~\ref{sec:applications}, we explore the advancements of Large Language Models (LLMs), particularly the GPT series, across various domains. 
Section~\ref{sec:synopsis} then summarizes the current perspectives and a range of research papers discussed in this review. In Section~\ref{sec:towards}, we present potential application areas of generalist AI, considering factors such as the psychotherapy process, scenarios, age groups, and specific psychological diseases. Finally, Section~\ref{sec:keypoints} highlights several key aspects that warrant focused attention for the future integration of AI in psychology, culminating in the conclusion in Section~\ref{sec:conclusion}.

\section{The potential of generalist AI models in psychology } \label{sec:generalist}
\begin{figure}[!hbtp]
\centering
\includegraphics[width=0.6\linewidth]{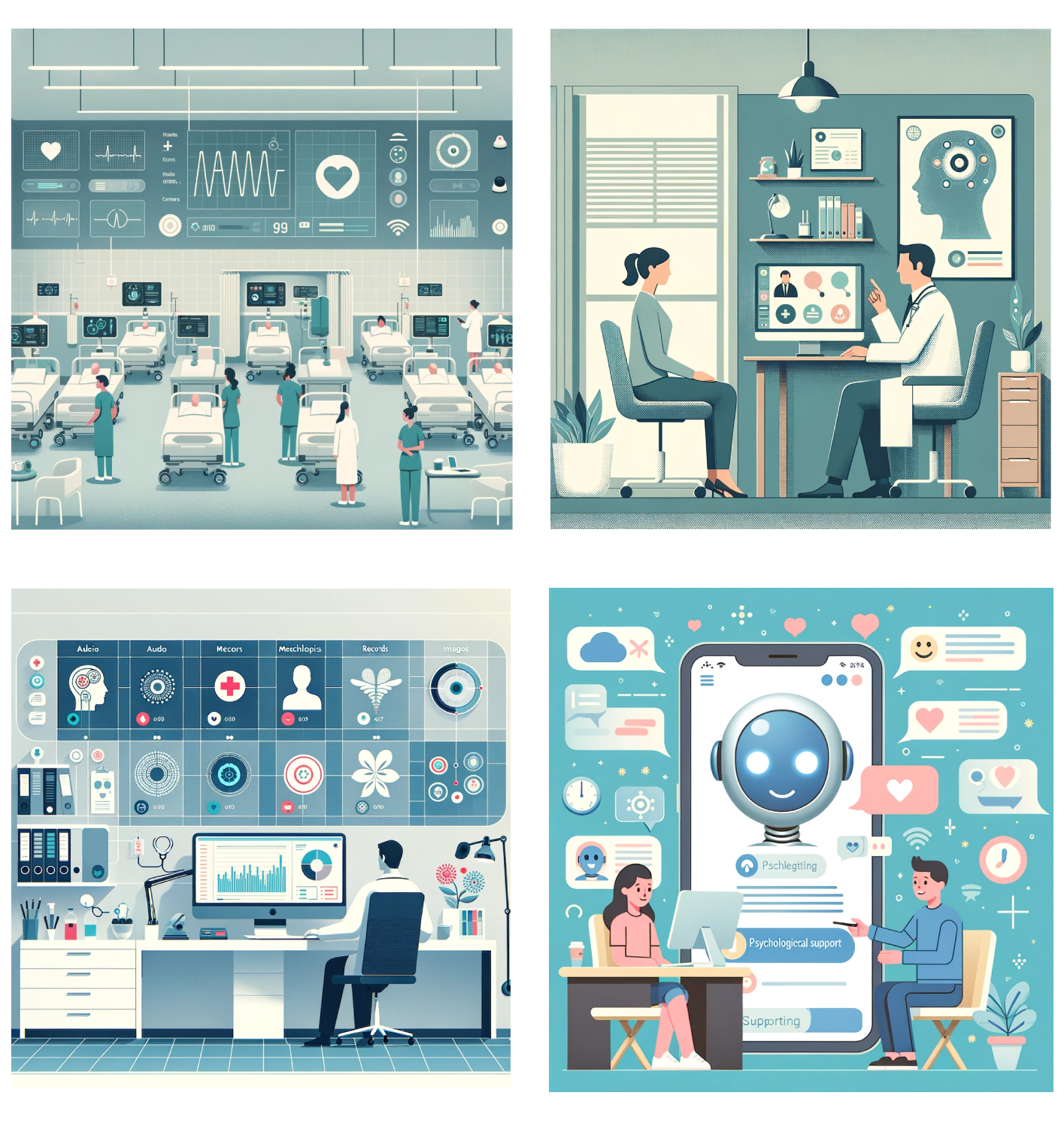}
\caption{Future applications of GPAI. (a) GPAI monitors patient health in the ward (b) GPAI assists doctors in psychological consultation (c) GPAI assists doctors in analyzing and diagnosing psychological data (d) GPAI supports people with negative emotions in social media.}
\label{fig:gpai_scenes}
\end{figure}

Generalist Psychology Artificial Intelligence (GPAI) holds promise for tackling a broader range of intricate tasks across multiple scenarios and demographic groups, surpassing the capabilities of current AI models. This innovation paves the way for specific tasks to be executed with minimal or even no labeling requirements. GPAI is designed to cater to the demands of diverse fields, ensuring a flexible and responsive user interaction. It encompasses several key functionalities. First and foremost, under ethical constraints, GPAI emphasizes appropriate monitoring: it seeks to diligently track user mental health across diverse situations, always prioritizing privacy and ensuring user transparency. Second, flexible interaction of data modalities: it exhibits the capability to interact with and integrate multiple data modalities. Even when users analyze complex data sets, GPAI can generate outputs that seamlessly combine these varied modalities. Third, advanced reasoning and prediction: GPAI leverages psychological and medical knowledge to conduct sophisticated reasoning. It can anticipate future psychological states and unearth novel psychological insights, enhancing understanding and potential interventions. Overall, GPAI aspires to be a multifaceted tool, adept at adapting to individual needs while offering advanced analytical capabilities in the realm of psychology. For a basic view of all prospective scenarios, refer to Figure~\ref{fig:gpai_scenes}.
\paragraph{Ethical psychological monitoring}
Given the fluctuating nature of human psychology and the intricate landscape of mental states, consistent real-time expert companionship and surveillance are challenging. Diseases like depression can manifest abruptly, leading, tragically, to suicides that might be prevented~\cite{hawton2013risk}. For instance, many depression sufferers, either on social media or in their daily lives, may not actively seek help~\cite{de2021identification}. However, they might reach out to a suicide prevent line~\cite{doupnik2020association} or express suicidal intentions online~\cite{chancellor2020methods} under some special topics as a cry for assistance. An individual who takes their own life has typically attempted suicide several times previously~\cite{hill2020association}. Herein lies the potential for generalist AI: introducing universally accessible suicide hotlines and ethically monitoring online platforms for potential risks. This can bridge the gap in expert availability and reduce suicide incidences. In psychiatric rehabilitation centers, GPAI can be employed to detect anomalies, including potential suicide attempts or other emergencies~\cite{haque2020illuminating}. In such applications, maintaining patient privacy remains of utmost importance~\cite{feng2022clinical}.

\paragraph{Multimodal human-AI collaboration}

GPAI should adeptly manage multi-modal data inputs provided by users and reciprocate with feedback in the desired format. The psychological domain commonly deals with textual medical records, visual data from social platforms, audio from interviews, tabulated psychological assessments, and psychoanalytic imaging data. Catering to diverse user needs might entail multiple data modality interactions. For instance, monitoring social media for potential risks requires GPAI to scrutinize both images and text, discerning varying suicide risks~\cite{gui2019cooperative, zhou2019multi}, and subsequently facilitating interventions or referrals at the optimal moment. In aiding doctors during psychological assessments or therapeutic sessions, GPAI should process real-time dialogues between patients and clinicians, offering intervention tactics and psychological profiles to enable more tailored treatments~\cite{fu2023enhancing}. Hence, GPAI's adaptability in handling multiple data modalities is indispensable~\cite{tu2023towards, moor2023foundation}.

\paragraph{Enhanced reasoning and predictive capabilities}

Equipped with a wealth of knowledge spanning domains like psychology and medicine, GPAI is poised to complement human expertise, deducing novel mental states and disease attributes based on established knowledge. For instance, the emergence of COVID-19 has impacted diverse demographic groups' psychological well-being. In these novel scenarios, GPAI can identify predominant psychological challenges faced by distinct groups—be it healthcare professionals~\cite{galbraith2021mental, liu2020mental}, adolescents~\cite{szlyk2020covid}, or the general public~\cite{li2020progression}—and proffer appropriate recommendations and interventions~\cite{sarker2021robotics}. GPAI holds promise in assisting experts in deciphering and modeling mental illness progression, such as forecasting potential disease trajectories using data on a patient's mood, familial, school or societal support structures, medication, and more, enriching our comprehension of these conditions. GPAI's predictive prowess, like estimating future suicide rates based on hotline interactions, can significantly aid experts in more precise interventions and suicide prevention~\cite{tong2023predictive}.

\section{Model evaluation} \label{sec:evaluation} 
% Model evaluation is crucial, and this type of research allows potential users to better comprehend the scope of competence and boundaries of large language models (LLMs), which can include some evaluation platforms~\cite{platform2023chatbotarena, platform2023openllm} such as CLEVA~\cite{li2023cleva} or other evaluation datasets~\cite{zheng2023lmsys, laskar2023systematic, wang2023decodingtrust} such as CHBias~\cite{zhao2023chbias}. This chapter primarily focuses on two aspects: the task-specific evaluations in Psychology~\ref{sec:evaluation:tasks} such as assessing the model's capabilities on datasets related to cognitive biases and suicide~\cite{qi2023evaluating}; and evaluation of the model's psychological aspects in certain scenarios, including examining the model's emotional intelligence and similar traits~\cite{wang2023emotional}. This kind of research can enhance our understanding of the specific performance of LLMs, which is important for simulating studies related to human subjects. Details are shown in the following text.  
Model evaluation is essential for understanding the capabilities and limitations of LLMs. It encompasses various evaluation platforms such as Chinese Language Models EVAluation Platform (CLEVA)~\cite{li2023cleva}, and datasets like CHBias~\cite{zhao2023chbias}, enabling potential users to grasp the models' scope of competence~\cite{platform2023chatbotarena, platform2023openllm, zheng2023lmsys, laskar2023systematic, wang2023decodingtrust}. This section focuses on two primary aspects: task-specific evaluations in Psychology, particularly in assessing models on datasets related to cognitive biases and suicide~\cite{qi2023evaluating}; and the evaluation of models' psychological characteristics in specific scenarios, such as their emotional intelligence~\cite{wang2023emotional}. This research enhances our understanding of LLMs' specific performance, crucial for studies simulating human subjects. Further details are discussed in the subsequent sections. The specifics of LLM evaluation are extensively covered in the literature~\cite{chang2023survey}.
% 已修改

\subsection{Task-specific evaluations in psychology} \label{sec:evaluation:tasks}
The performance of LLMs in the field of psychology can be assessed through various task-based validation methods with specific criteria for evaluation. These tasks could include emotion recognition or emotional-awareness, and mental health classification such as depression detection or suicidal risk detection.

Recently, researchers indicated that the existing uses of AI for the field of mental health are limited in their capabilities in emotional-awareness~\cite{pham2022artificial}. Emotional-awareness was assessed using the level of emotional awareness scale (LEAS) in twenty different scenarios, comparing GPT-3.5's performance with general population norms~\cite{elyoseph2023chatgpt, nandrino2013level}. A subsequent evaluation assessed GPT-3.5's progress in emotional-awareness and involved independent licensed psychologists evaluating the contextual appropriateness of GPT-3.5's emotional-awareness responses, reaching almost the maximum LEAS score (Z score=4.26), with an exceptional accuracy level of 9.7/10.
Wake N, et al.~\cite{wake2023bias} investigated GPT-3.5's ability in emotion recognition from text, essential for applications like interactive chatbots and mental health analysis. While prior studies demonstrated GPT-3.5's proficiency in positive or negative sentiment analysis~\cite{amin2023will, kocon2023chatgpt}, its nuanced emotion recognition remained unexplored. Experiments revealed reproducible performance with improvements through fine-tuning. However, performance inconsistencies across emotion labels and datasets suggested inherent instability and potential bias. The choice of dataset and emotion labels significantly influenced GPT-3.5's emotion recognition. The study emphasizes the importance of dataset selection and fine-tuning, laying the foundation for improved emotion analysis integration in GPT-3.5 applications.
Furthermore, some researches have explored LLMs' capacities of identifying mental health issues, which included stress~\cite{lamichhane2023evaluation}, depression~\cite{lamichhane2023evaluation}, cognitive distortions~\cite{qi2023evaluating}, and suicidal risk~\cite{qi2023evaluating, lamichhane2023evaluation, levkovich2023suicide}. Using supervised learning as a baseline, a study experimented two pivotal tasks: suicidal risk and cognitive distortion identification on Chinese social media platforms through three distinct strategies: zero-shot, few-shot prompt, and fine-tuning~\cite{qi2023evaluating}. The findings revealed a noticeable performance gap between LLMs and traditional supervised learning approaches, primarily due to the models' inability to fully grasp subtle categories. Notably, while GPT-4 outperformed other versions in multiple scenarios, GPT-3.5 exhibited significant enhancement in suicide risk classification after fine-tuning.
Lamichhane's study evaluate the performance of LLM-based ChatGPT in various mental health classification tasks, reporting F1 scores of 0.73 for stress detection, 0.86 for depression detection, and 0.37 for suicidality detection~\cite{lamichhane2023evaluation}. Levkovich I and Elyoseph Z~\cite{levkovich2023suicide} focuses on ChatGPT's ability in suicide risk assessment, particularly analyzing two discernible factors: perceived burdensomeness and thwarted belongingness. Their study includes comparisons between assessments conducted by GPT-3.5, GPT-4 and mental health professionals. The findings suggest that GPT-4's assessments of suicide attempt likelihood are in close agreement with those made by mental health professionals. This reveals its accuracy in identifying suicidal ideation. However, its tendency to overestimate psychological pain (psychache) points to the necessity for additional research. In comparison, GPT-3.5 often underestimates suicide risk, highlighting concerns about its precision, particularly in critical cases. These results underscore the potential of LLMs in mental health settings, laying the groundwork for their continued improvement and wider use.

\subsection{LLM performance in psychological contexts} \label{sec:evaluation:perform}
Recent studies have started to assess the exploratory abilities of LLMs in psychological context. While this kind of evaluation often lack quantfiable criteria, but can let people understand how it performs. This exploration is particularly crucial for exploring situations unknown to humans or unclear to define.
Ullman investigated the performance of LLMs in reasoning tasks related to intuitive psychology~\cite{ullman2023large}.This study highlists that several recent tasks and benchmarks assessing the reasoning ability of LLMs have primarily focused on belief attribution in Theory-of-Mind tasks~\cite{ullman2023large}. 
He et al.~\cite{he2023homophily} have initiated research on using LLM as the driving core of social bots. They employed tasks such as profile completion, social behavior decision-making, and social content generation to explore models' capabilities. Furthermore, benchmarks in the domains of planning and explanation generation were introduced to test the limit of LLMs' abilities in reference~\cite{collins2022structured}. To enhance problem solving and human-like reasoning in LLMs, a hybrid Parse-and-Solve model was developed to integrate distributional LLMs with a structured symbolic reasoning module.
Experiments have been conducted to testify LLMs' communication and emotional abilities. A new psychometric assessment focusing on emotion understanding was developed to assess Emotional Intelligence, which includes emotion recognition, interpretation, and understanding, all crucial for effective communication and social interactions~\cite{wang2023emotional}. This test is applicable to both humans and LLMs and involves interpreting complex emotions in realistic scenarios. Additionally, an opinion network dynamics model that incorporates LLMs' opinions, individuals' cognitive acceptability, and usage strategies, was created to explore its capacity in effective demand-oriented opinion network interventions~\cite{li2023quantifying}. 

\section{Progress and Applications of LLM in Psychology} \label{sec:applications}
In this section, we introduce the applications of LLM in three scenarios: digital media platforms, clinical nursing care practice and community. We also discuss its progress in exploring psychological theories and data generation.

\subsection{Digital and social media platforms} \label{sec:app:social_media}

\subsubsection{Sentiment detection and analysis} \label{sec:app:social_media:sentiment}
The detection of psychological constructs, mental state and suicide risk through social media content has become increasingly important for early diagnosis and prevention of adverse consequences related to user's psychological states~\cite{rissola2021survey, coppersmith2018natural}. 
Qin et al.~\cite{qin2023read} developed a system that provides precise diagnoses and evidence based insights. This system analyze user's mental state through social media conversations. This system offers tailored recommendataions to promote well-being and integrates advanced techniques with expert criteria to enable informed decisions. The experiments demonstrate that the system achieves the best performance across various settings, including full data setting, few-shot setting, zero-shot setting, independent-identical-distribution (IID) setting, and out-of-distribution (OOD) setting. Notably, under the IID setting, the system consistently attains optimal performance across various data configurations and metrics. Rathje et al.~\cite{rathje2023gpt} explored GPT-3.5's capability for automated psychological text analysis in various languages, finding its accuracy in detecting  psychological constructs like sentiment, discrete emotions, and offensiveness to be satisfactory and superior to many current automated text analysis methods. 
% Furthermore, it is possible that LLMs are also helpful in detecting suicide risk. A study conducted in 2018 utilized natural language processing and machine learning techniques, specifically deep learning, to detect quantifiable signals of suicide attempt and described the design of an automated system for estimating suicide risk~\cite{coppersmith2018natural}. 
LLMs like GPT-3.5 are valuable for screening suicide risk on social media.
Qi et al.~\cite{qi2023evaluating} introduced a comprehensive benchmark for evaluating LLMs on cognitive distortions or suicidal risk. This benchmark assesses the LLM's abilities in detecting these concerns using social media data. As task complexity increases, the performance of LLMs tends to decline. While fine-tuning LLMs can enhance their effectiveness in relatively straightforward task like suicide classification, this approach falls short in more complex cognitive distortion task. This indicates a need for further research into advanced fine-tuning mechanisms for LLMs.
% Since recognizing signs of suicide risk or other mental health problems in a timely manner is crucial for effective interventions, recently, a comprehensive benchmark using supervised learning and LLMs were proposed by Qi et al.~\cite{qi2023evaluating} to recognize real-time signs of cognitive distortions or suicidal tendencies. This research assessed the capabilities of LLMs using social media data.

\subsubsection{Intelligent Q\&A chatbots} \label{sec:app:social_media:chatbots}
Q\&A Chatbots powered by LLMs have increasingly been utilized in mental health support~\cite{ma2023understanding}. 
Cabrera et al.~\cite{cabrera2023ethical} recommended the use of LLM based Q\&A chatbots, specifically tailored for mental health to complement the care provided by human professionals. A qualitative analysis of 120 posts, including 2917 user comments showed that the Replika app offers on-demand, non-judgmental support, boosts user confidence, and aiding self-discovery~\cite{ma2023understanding}. 
In a patient care context, Chen et al.~\cite{chen2023llm} demonstrated the feasibility of utilizing a ChatGPT-driven chatbot in mental health scenarios. The chatbot, capable of simulating experts or patients, showed a high level of similarity in its interactions. Cai et al.~\cite{cai2023paniniqa} developed PaniniQA, a system designed to improve patient understanding of their discharge instructions through interactive question answering. They claimed that the sytem is capable of identifying important clinical content and effectively generating patient-specific educational questions.
Despite the benefits, there are still some challenges in LLM based Q\&A Chatbots. 
Ma et al.~\cite{ma2023understanding} identified issues such as difficulty in filtering harmful content, maintaining consistent communication, managing the retention of new information, and addressing user overdependence. Fu et al.~\cite{fu2023enhancing} employed a vector database to detect content produced by LLMs in online psychological counseling support system, thereby preventing the generation of harmful opinions. Additionally, there is a concern that social stigma associated with using these chatbots could further isolate users~\cite{ma2023understanding}. 
Several studies strongly assert that future researchers and designers must conduct a comprehensive evaluation of LLMs in mental well-being support. This is essential to ensure the responsible use and proper regulation at both national and international levels~\cite{ma2023understanding, cabrera2023ethical}.

\subsubsection{Online psychological support tools} \label{sec:app:social_media:support}
Several studies have demonstrated that LLMs like GPT-3.5 or GPT-4 can effectively assist human in online psychological counseling and answering professional questions with increased empathy and acceptable feasibility~\cite{ayers2023comparing, fu2023enhancing, chen2023llm}. 

Ayers et al.~\cite{ayers2023comparing} conducted experiments to compare the response of physicians and chatbots to patient questions on a public social media platform. The results showed that responses generated by chatbots were preferred and rated significantly higher in terms of both quality and empathy. This finding indicates that AI assistants could be valuable in assisting physicians in formulating responses to patient queries, potentially enhancing the effectiveness and emotional resonance of online mental consultations. 
In light of the increasing negative emotions especially concerning suicidal intentions expressed on social media, there is a critical need of trained psychological counselors~\cite{fu2023enhancing}. However, the shortage of trained professional counselors has prompted the necessity to involve non-professionals or volunteers in providing online psychological support. Fu et al.~\cite{fu2023enhancing} proposed a novel model based on LLMs to empower non-professional counselors to offer effective psychological interventions online. An evaluation involving ten professional counselors demonstrated the system's efficacy in analyzing patients' issues and offering professional-level recommendations. The utilization of such models can effectively bridge the gap between the demand for psychological counseling and the limited availability of professional resources. 
Cacers et al.~\cite{caceres2023wmgpt} introduced Well-Mind ChatGPT (WMGPT), a system offering 24/7 mental health counseling services which utilized a fine-tuned ChatGPT model. This system presents a promising solution by offering accessible, personalized, and timely mental health support. It effectively overcomes the limitations of traditional counseling methods and enhancing overall well-being. 
Lai et al.~\cite{lai2023psy} proposed Psy-LLM framework, an AI system employs LLMs to enhance question-answering capabilities in online psychological consultation. The effectiveness of this system was evaluated using metrics such as perplexity and human assessments, demonstrating the capability in generating coherent and relevant answers. 
These studies emphasized the potential of LLMs to enhance mental health support in online psychological consultation.

\subsection{Clinical and care settings} \label{sec:app:clinical}

\subsubsection{Mental interview, psychological counselor assistance} \label{sec:app:clinical:support}
Currently, there is a contradiction between the increasing number of individuals experiencing emotional distress and the limited number of professional counselors, coupled with lengthy training periods and high costs. LLMs can play a significant role in assisting psychological counseling scenarios. 
Eshghie et al.~\cite{eshghie2023chatgpt} suggested using ChatGPT as an adjunct in psychotherapy, where it could act as a tool for collecting patient information, provide companionship to patients between therapy sessions, and organize information for therapists to streamline treatment processes. GPT-3.5 can participate in positive conversations and provide validation or coping strategies, which can help therapists engage in the counselling. Chen S et al.~\cite{chen2023llm} utilized ChatGPT to simulate psychiatrist and patient, collaborating with psychiatrists to develop the dialogue system aligned with real-world scenarios.  Furthermore, LLMs have demonstrated exceptional performance in online counseling. Research indicates that LLMs can accurately analyze patient issues and provide professional-level recommendations, thereby augmenting support for non-professional counselors~\cite{fu2023enhancing}. 

\subsubsection{Wearable devices and behavior monitoring} \label{sec:app:clinical:wearable}
LLMs excel are adept at grasping complex concepts, yet real-world health applications require integration with numerical data.
Liu et al.~\cite{liu2023large} showed that, LLMs can effectively analyze physiological and behavioral time-series data with minimal tuning. This capability extends to tasks such as cardiac signal analysis, physical activity recognition, metabolic calculation, and stress estimation using data from wearable and medical sensors.

% \subsubsection{Ambient Intelligence} \label{sec:app:clinical:ambient}

\subsection{Community-based psychological screening tools} \label{sec:app:community}

\subsubsection{Focused on children and adolescents} \label{sec:app:community:children}
LLMs offer notable advantages for mental health support among children and adolescents. Specifically, GPTs show notable potential in offering initial assistance to young adults facing mental health challenges~\cite{aminah2023considering}. Dosovitsky G and Bunge E~\cite{dosovitsky2023development} explored the use of chatbots for mental health among adolescents, an area that has been less studied in this context compared to adults. Twenty-three participants aged 13–18 tested a chatbot designed to educate about depression, teach behavioral activation, and helping in addressing negative thoughts. The study revealed moderate level of satisfaction among participants with over half providing positive feedback. This suggests that mental health chatbots are generally well-received by adolescents and can be a valuable tool in addressing their mental health needs.
While recognizing the benefits, it is crucial to carefully address the limitations of LLMs which necessitate more thorough investigation~\cite{aditama2023epistemic, mardikaningsih2023risk}. Aditama et al.~\cite{aditama2023epistemic} identified that the LLM's limitation including epistemic injustice like interpretive value and an ethically flawed affective value which may lead to a diminished effectiveness. More seriously, relying solely on LLMs for mental health support without professional guidance might overlook essential assessments, monitoring, and treatment, potentially increasing the risk of suicide attempt~\cite{mardikaningsih2023risk}. LLM applications for children should prioritize safety, particularly within an ethical framework.

\subsubsection{Educational settings and students} \label{sec:app:community:education}
Researchers are increasingly exploring the use of LLMs in mental health education and as tools for student social engagement. As an interactive chatbot powered by GPT-3.5 can be effective in educating students about psychological impacts of sexual violence~\cite{folastri2023chatgpt}.  Its interactive and accessible nature makes it a valuable tool for disseminating information, aiding in the prevention of sexual violence, and supporting treatment efforts. Moreover, integrating GPT-3.5 into educational curricula can raise awareness and assist affected students. 
LLMs may also serve as alternatives for students' interpersonal relationships. Maples et al.~\cite{maples2023loneliness} conducted a survey with over 1000 student users of the intelligent social agent, Replika. They found that these participants, despite being lonelier than typical students, perceived high social support from Replika.  The chatbot, perceived by users as a friend, therapist, or intellectual mirror, could play a significant role in reducing feelings of loneliness and suicidal ideation among the students.

\subsection{Advancing psychological theories} \label{sec:app:theories}

\subsubsection{Simulating social networks} \label{sec:app:theories:social}
The use of LLMs can significantly contribute to the study of simulated social networks, providing researchers with valuable tools to observe and explore various psychological theories. He et al.~\cite{he2023homophily} conducted a study exploring the behavior of LLMs within a simulated social network. They designed a Twitter-like platform called Chirper that populated only by LLMs. Their analysis revealed evidence of self-organized homophily in the system, demonstrating patterns that resemble human social tendencies. Although the content generated by chatbots is generally more generic compared by human, this observation highlights the potential for developing LLM-driven Agent-Based Models. Such models could enhance the comprehensive understanding of human social dynamics in social scientific research.
Gao et al.~\cite{gao2023s} leveraged the abilities of LLMs to develop the Social Network Simulation System. This system, employing agent-based simulation techniques, effectively replicated authentic human behavior, capturing nuances in emotions, attitudes, and interactions. 
Park et al.~\cite{park2022social} introduced a novel prototyping method called social simulacra. This technique employs LLMs to simulate the interactions of diverse community members. It is based on a designer's specifications about the community's design goals, rules, and member personas. 
The evaluation of both studies have shown promising accuracy, marking a signifying step forward in the domain of LLM-based social network simulation. 

Besides, based on certain theories, LLMs have been evaluated for specific capabilities. A recent study investigated the social intelligence and Theory of Mind (ToM) capabilities of GPT-3. This involves the ability to understand various mental states, intentions, and reactions, which is crucial for effectively navigating everyday social interactions~\cite{sap2022neural}. The research assessed GPT-3's performance on SocialIQa and ToMi tasks, focusing on its comprehension of social interactions and ability to deduce mental states. The findings revealed that GPT-3's accuracy in these areas is below human levels.

\subsubsection{Modeling individual agents} \label{sec:app:theories:agents}
Another emerging application of LLMs is in simulating specific populations, aiding research focused on these demographics.Various studies have evaluated the accuracy of these simulations. The methods used include conducting Turing Experiments~\cite{aher2023using}, comparing the simulated responses with those from human samples ~\cite{argyle2023out}, and recruiting real psychiatrists or patients in diagnostic conversations with the chatbots~\cite{chen2023llm}. These approaches help in assessing the realism and relevance of LLM simulations in representing actual populations.
Chen et al.~\cite{chen2023llm} and Aher et al.~\cite{aher2023using} demonstrated the potential of LLMs in accurately simulating diverse human interactions for various research and practical applications. 
% Psychiatrist and patient included, LLMs are proposed as effective proxies for specific human sub-populations, enabling the generation of realistic social interactions and interactions with desired properties based on provided prompts~\cite{chen2023llm, aher2023using}. 

\subsubsection{Simulating environments and contexts} \label{sec:app:theories:environments}
Currently, interpersonal interactions such as transactions, role reversal, as well as interactive scenarios like classrooms and communities simulated by LLMs are emerging. 

LLMs extend beyond simulating interactions between two individuals; they are equally adept at simulating interactions among multiple individuals in settings such as community or educational environments. Park et al.~\cite{park2023generative} introduced generative agents, computational entities that mimic genuine human behavior. These agents are capable of possessing life experiences, making autonomous decisions, and engaging in realistic interactions. They can form opinions and participate in conversations. When implemented in an interactive sandbox environment, these agents allow user interaction through natural language. This represents a significant advancement in the creation of believable simulations of human behavior, achieved by integrating LLMs with interactive computational agents. 
Shi et al.~\cite{jinxin2023cgmi} successfully replicated realistic human interactions in various scenarios, particularly focus on classroom environment. They utilized a framework named Configurable General Multi-Agent Interaction, based on LLMs. The experiments conducted with this framework showcased its capability to closely mimic real classroom dynamics, including teaching methods, curriculum, and student performance.
% Furthermore, LLMs can also simulate environments to assess agents' ability to mimic various roles presented by the other agent.
% Kovavc et al.~\cite{kovavc2023socialai} proposed a LLMs-driven SocialAI school, a tool that designed to customizable procedurally generated simulating environments for experiments related to socio-cognitive abilities of LLMs. Models' capabilities like role-reversal was examined and findings suggest a promising avenue for researches on directing the agent's attention to the partner's role. 
Furthermore, LLMs have the capability to simulate environments, assessing the agents' proficiency in mimicking diverse roles presented by another agent. Kovavc et al.~\cite{kovavc2023socialai} introduced a SocialAI school driven by LLMs, a tool designed for customizable, procedurally generated environments to facilitate experiments related to the socio-cognitive abilities of LLMs. This platform enabled the examination of model capabilities such as role-reversal. The findings indicate a promising research direction in guiding an agent's attention towards understanding and adopting the partner's role.

\subsubsection{Applications and innovations in psychological theory} \label{sec:app:theories:innovations}
As technologies progress, the application of LLMs in field such as understanding mentalizing abilities and personality structures is growing, particularly focusing on the exploration of human emotions on a theoretical bases. Despite the fundamental differences between machine intelligence and human emotions, LLMs have demonstrated potential in adopting various personalities~\cite{ratican2023six}. 
Hadar-Shoval et al.~\cite{hadar2023plasticity} explored the ability of GPT-3.5 to generate mentalizing-like responses tailored to specific personality structures, such as Borderline Personality Disorder and Schizoid Personality Disorder. While the responses aligned with clinical knowledge, concerns about potential stigmas and biases related to mental diagnoses emphasized the necessity for responsible development in chatbot-based mental health interventions. Binz and Schulz~\cite{binz2023turning} transformed LLMs into cognitive models by fine-tuning with psychological experiment data. This resulted in accurately representing human behavior and outperforming traditional cognitive models in decision-making domains. 
Ullman et al.~\cite{ullman2023large} explored the intuitive psychology in LLMs for belief attribution in theory-of-mind tasks revealing both successes and failures. The research emphasized skepticism about zero-hypothesis in model evaluation and highlighted the importance of prioritizing exceptional failure cases over average success rates. In conclusion, LLMs' adaptability, even in unseen tasks, suggested their potential as generalist models, paving new pathways in cognitive psychology, behavioral sciences and intuitive psychology~\cite{binz2023turning, ullman2023large}. 

\subsection{Data Generation techniques} \label{sec:app:data}
Language and cognition research extensively relies on large-scale psycho-linguistic datasets, which include judgments on lexical properties such as concreteness and age of acquisition~\cite{trott2023can}. These datasets are essential for standardizing experimental stimuli, revealing empirical relationships within lexicons, training deep learning models, and serving a range of research objectives. However, the economic and time constraints associated with data collection make acquiring such large scale dataset increasingly challenging. Trott~\cite{trott2023can} investigated the potential of utilizing advancements in LLMs, particularly GPT-4, to facilitate the creation of extensive psycho-linguistic datasets in English. This research involved a comparison between semantic judgments generated by LLMs and those made by human, revealing positive correlations. In some instances, LLM-generated judgements even exceeded human inter-annotator agreement. The study also examined the systematic differences between norms generated by LLMs and humans, providing further insights into the performance of LLMs in this context.

\section{Synopsis of the viewpoints in literature}\label{sec:synopsis}
In this section, we provide a summary of the viewpoints expressed in current reviews and opinion papers regarding the application of LLMs in the medical field and associated domains.
% 请在这章节总结各个文献中的观点，根据情况表格描述或者文献描述，乃至思维导图描述。

\subsection{Potential application}
In summary, these references~\cite{he2023survey, min2023recent, arora2023promise, demszky2023using, zhong2023artificial, thirunavukarasu2023large} highlight the wide-ranging applications of LLMs in healthcare, including tasks like medical question answering, generation of comprehensive assessments, clinical conversations, and medical research. The works of Demszky et al.~\cite{demszky2023using} and Zhong et al.~\cite{zhong2023artificial} discuss LLMs' potential in measurement and experimentation, particularly in psychiatric research within psychology subfields. Zhong et al.~\cite{zhong2023artificial} further suggest that LLMs can enhance diagnostic accuracy and personalized treatment in neuropsychiatry. Thirunavukarasu et al.~\cite{thirunavukarasu2023large} underscore the necessity of addressing ethical and technical challenges in LLM deployment. Moor et al.~\cite{moor2023foundation} emphasize the need for generalist medical artificial intelligence (GMAI) to adapt flexibly to new tasks and data patterns. Cabrera et al.~\cite{cabrera2023ethical} focus on LLMs' potential in improving mental health counseling, patient care, diagnostic accuracy, and treatment planning. Mao et al.~\cite{mao2023gpteval} observe that LLMs' task performance in specific fields still falls short of expert models. Lastly, Kaddour et al.~\cite{kaddour2023challenges} stresses the importance of LLM-based toxicity detection tools in mitigating harmful content output.

\subsection{Bias and fairness issues}
Overall, the references \cite{he2023survey, demszky2023using, moor2023foundation, cabrera2023ethical, mao2023gpteval, zhao2023survey, kaddour2023challenges, arora2023promise} underscores fairness as a critical issue in the utilization of LLM and natural language processing (NLP), highlighting potential biases that may lead to inconsistencies, unequal treatment, and discrimination. These biases in LLMs can manifest as prejudiced sentiments based on gender, race, religion, or political beliefs, potentially affecting healthcare outcomes, such as the underdiagnosis in marginalized communities. Further, the references~\cite{keeling2023algorithmic, thirunavukarasu2023large, min2023recent, kaddour2023challenges} notes that biases in LLMs may arise from perspective biases in the training data sourced from internet text corpora, thus raising ethical concerns. Keeling et al.~\cite{keeling2023algorithmic} suggest that ethical nudges, though not a complete solution, can mitigate bias when combined with appropriate evaluative measures. They emphasize the importance of inclusive and open discussions on who fine-tunes these models and the standards applied. Demszky et al.~\cite{demszky2023using} argue that while addressing bias is crucial, examining it can also provide insights into the dominant culture of the training data. Moor et al.~\cite{moor2023foundation} recommend ongoing audit monitoring and prize competitions to encourage scrutiny and bias identification. Additionally, references~\cite{mao2023gpteval, zhao2023survey} highlights the challenges LLMs' generative capabilities pose to fact evaluation. Zhao et al.~\cite{zhao2023survey} raise concerns regarding copyright and personal information leakage. Arora et al.~\cite{arora2023promise} notes the impact of generated inflammatory content on mental health, while Zhong et al.~\cite{zhong2023artificial} point out the need for reliability, accuracy, and transparency in LLMs, particularly in their application to mental illness, emphasizing the focus on ethical issues.

\subsection{Moderation approach}
He et al.~\cite{he2023survey} highlight a range of strategies for mitigating bias in LLMs, including the integration of specific anti-discrimination components and modules, alongside prompt engineering and instruction fine-tuning to promote fairness. Keeling et al.~\cite{keeling2023algorithmic} argue for the re-evaluation of biases in generalist models, which may differ from previously held assumptions about systematic biases. Demszky et al.~\cite{demszky2023using} emphasize the need for human expert oversight in LLM interactions, particularly when dealing with vulnerable groups. Thirunavukarasu et al.~\cite{thirunavukarasu2023large} note that adversarial prompts can exploit vulnerabilities in LLMs, leading to toxic outputs.
These references~\cite{moor2023foundation, arora2023promise} stresses that GMAI models, after accessing extensive patient data, may face privacy issues. It's essential to minimize the risk of sensitive data exposure and to review potential attacks on the model effectively. Cabrera et al.~\cite{cabrera2023ethical} assert that interdisciplinary research involving diverse talents is crucial to avoid bias.
Mao et al.~\cite{mao2023gpteval} suggest using Reinforcement Learning with Human Feedback (RLHF) to reduce bias and toxic responses in LLMs, while cautioning against system gaming and aggressive reward loops that could mislead the training process. Zhao et al.~\cite{zhao2023survey} state that adversarial methods can effectively explore LLMs to prevent harmful outputs.
Singh et al.~\cite{singh2023artificial} commend the American Psychiatric Association (APA) for creating a Digital Psychiatry Task Force and introducing an application evaluation model to assess the efficacy, safety, and tolerability of digital applications. The references~\cite{zhong2023artificial, singh2023artificial} underlines the importance of standardization, monitoring, and guidance to address potential safety concerns. Zhong et al.~\cite{zhong2023artificial} highlight the need for transparency and accountability mechanisms in these technologies.

\subsection{Privacy Issues}
The references~\cite{he2023survey, keeling2023algorithmic, demszky2023using, thirunavukarasu2023large, moor2023foundation, cabrera2023ethical, zhao2023survey} underscores the significance of privacy in healthcare applications, emphasizing the need for careful evaluation and regulation. He et al.~\cite{he2023survey} suggest that Federated learning models hold promise for protecting patient privacy while enabling large-scale evaluation of LLMs. Arora et al.~\cite{arora2023promise} acknowledge emerging solutions like federated learning and synthetic data but note uncertainties regarding their suitability for LLMs.
Demszky et al.~\cite{demszky2023using} and Singh et al.~\cite{singh2023artificial} argue that clear data privacy and protection processes are crucial for ethical practices in psychological research. Moor et al.~\cite{moor2023foundation} highlight the risks to patient privacy posed by GMAI models, which acquire rich patient characteristics. The potential exposure of sensitive patient data in training datasets and the risk of sensitive information being revealed through prompt attacks are major concerns. They advocate for de-identification and restricting information collection to mitigate these privacy risks.
Conversely, Thirunavukarasu et al.~\cite{thirunavukarasu2023large} point out a conflict: while privacy protection is critical, limiting patient identifiable data in model prompts could hamper deployment opportunities. This situation presents a potential conflict between personal identity and legal rights, underscoring the complexity of balancing privacy concerns with the effectiveness of LLMs in healthcare.

\subsection{Monitoring and evaluation}
The references~\cite{he2023survey, keeling2023algorithmic, thirunavukarasu2023large, moor2023foundation, cabrera2023ethical, arora2023promise, chang2023survey, kaddour2023challenges, zhong2023artificial} emphasizes the necessity of continuous monitoring and dynamic evaluation of LLMs in the medical field, focusing on aspects such as accuracy, fairness, transparency, and adherence to ethical standards. Keeling et al.~\cite{keeling2023algorithmic} highlight the need for the development and integration of bias mitigation techniques into regulatory policies, suggesting that evaluations should consider not just performance but also clinical semantic biases.
Demszky et al.~\cite{demszky2023using} advocate for clear processes to quantify and reduce bias, emphasizing the importance of investments in datasets, the standardization of performance benchmarks, and shared infrastructure to address ethical concerns. Thirunavukarasu et al.~\cite{thirunavukarasu2023large} note the challenges in assessing LLM performance on clinical tasks, particularly regarding their impact on mortality and morbidity. They suggest that randomized controlled trials are essential, but finding appropriate benchmarks and endpoints is complex.
Cabrera et al.~\cite{cabrera2023ethical} call for the monitoring and evaluation of chatbots' impact on mental health outcomes to ensure the quality, safety, and effectiveness of these interventions. The references~\cite{mao2023gpteval, min2023recent} points out that current evaluations, heavily reliant on prompt engineering and standard datasets, may be unreliable. It also raises the question of how to fairly compare the performance of expert systems and LLMs in professional fields.
Zhao et al.~\cite{zhao2023survey} state that while manual evaluation can accurately measure LLM performance in real-life scenarios, it is time-consuming and not scalable. They suggest an efficient and scalable model-based evaluation. Chang et al.~\cite{chang2023survey} advocate for a trustworthy and unified evaluation system to provide comprehensive guidance and analysis for future research. Finally, the references~\cite{kaddour2023challenges, zhong2023artificial} underscores the importance of detection tools, such as those for identifying toxicity and deceptive outputs, in the ongoing development and application of LLMs.

\section{Towards generalist AI in psychology} \label{sec:towards}
\begin{figure}[!hbtp]
\centering
\includegraphics[width=0.6\linewidth]{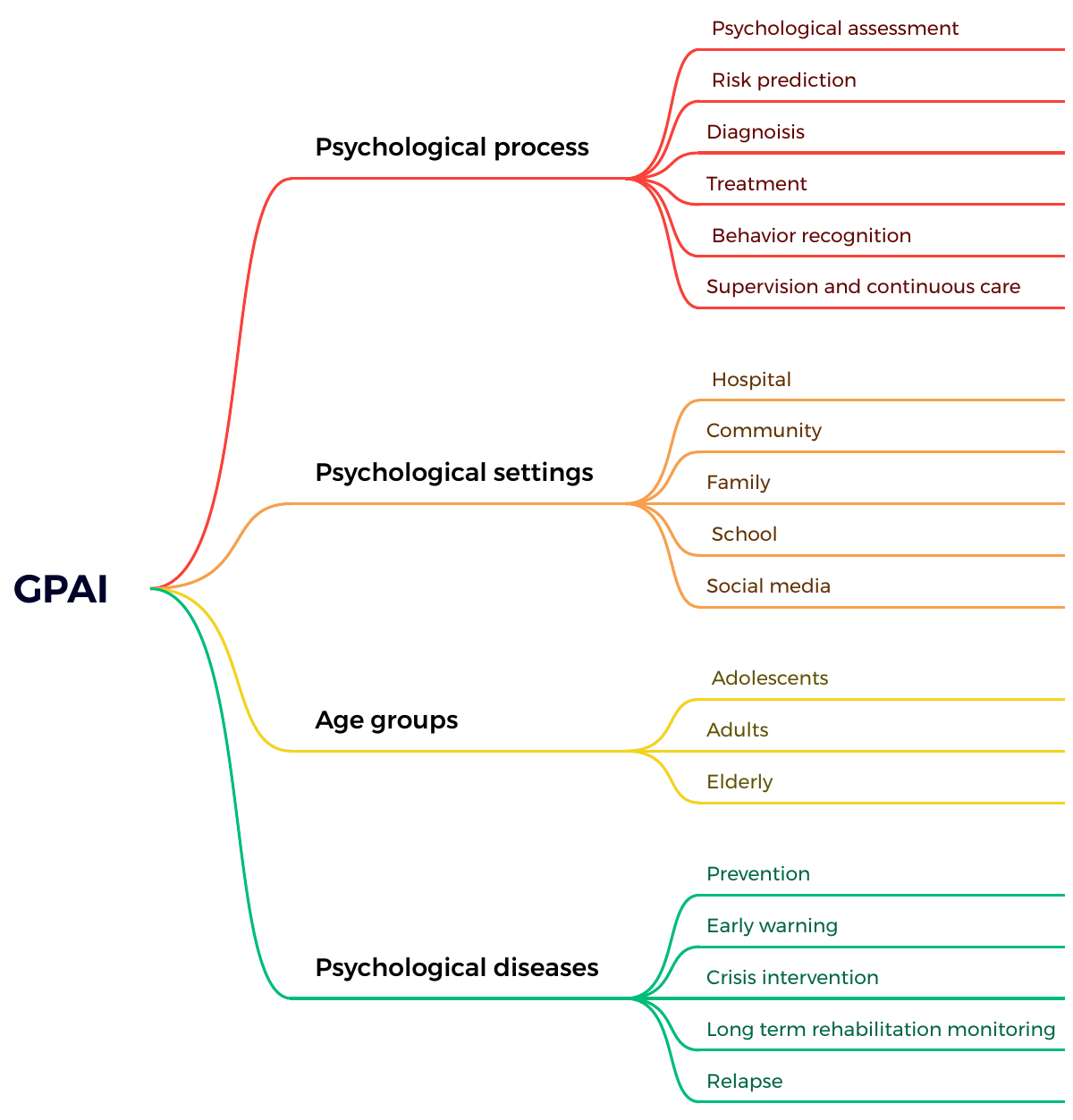}
\caption{Applications of GPAI in various aspects of psychology.}
\label{fig:gpai_application}
\end{figure}

\begin{figure}[!hbtp]
    \centering
    \includegraphics[width=0.9\linewidth]{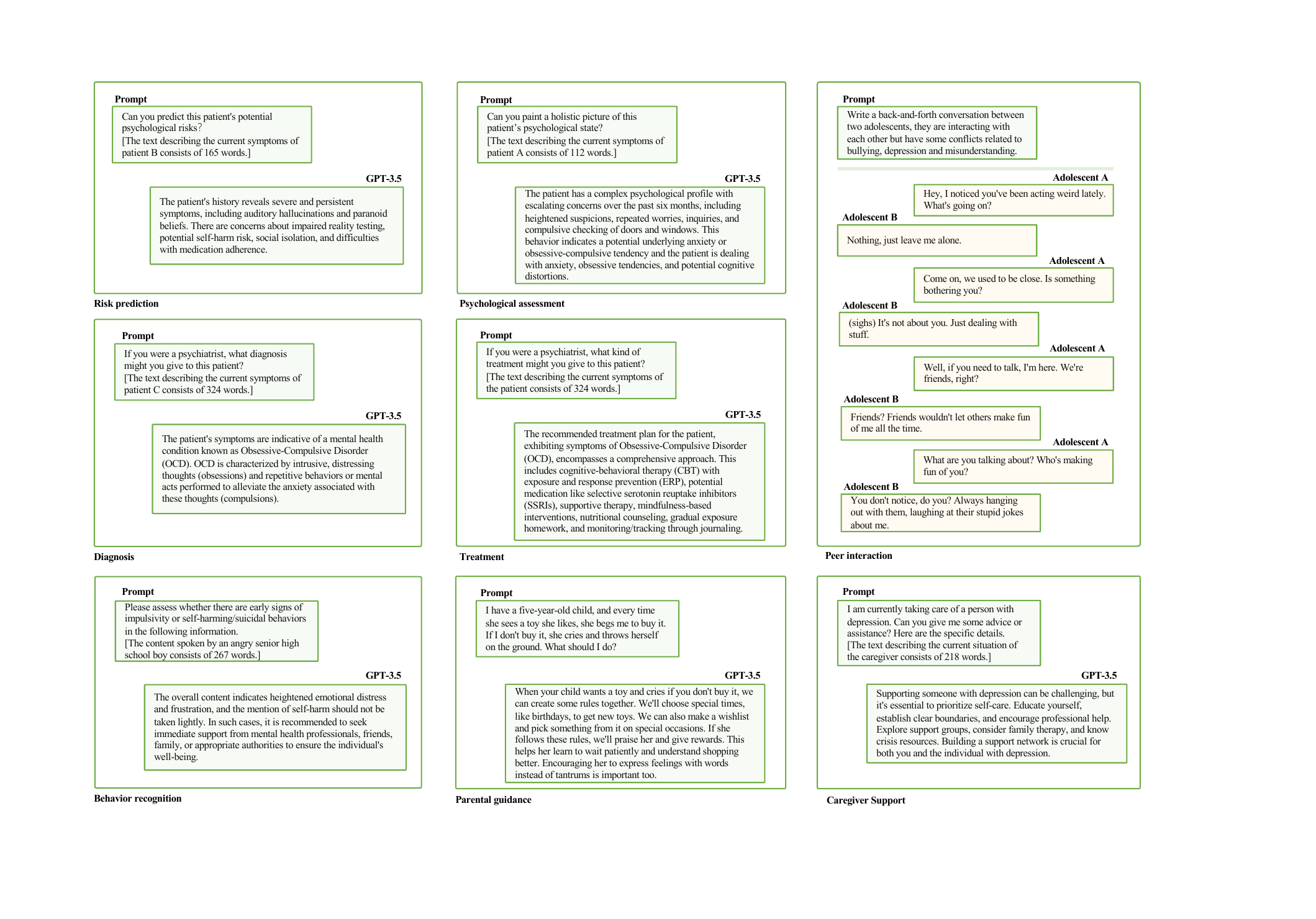}
    \caption{Experimental psychological scenarios utilizing the large language model (GPT-3.5) for testing purposes.}
    \label{fig:gpt_example}
\end{figure}

In this section, we explore the diverse potential applications of generalist AI in the field of psychology. The various application directions are illustrated in Figure~\ref{fig:gpai_application}. 
Additionally, Figure~\ref{fig:gpt_example} employs GPT-3.5 as an exemplary model to showcase the latest developments in AI. While GPT-3.5 does not exemplify generalist AI, it effectively highlights the significant strides achieved in the field to date. The examples depicted in the figure correspond to the content discussed in the subsequent sections.

% For example, as the dialogue shown in Figure~\ref{fig:gpai_application}, GAI has the capability to accomplish applications in various fields of psychology. In the future, it will probably possess the capability of multimodal input and output, allowing integration of various modalities such as text, video, audio, etc., based on the user's requirements. It generates results that are both professional and targeted, and can provide outputs in different modalities according to real-world context and needs.

\subsection{Generalist AI in psychotherapy process}
\paragraph{Psychological assessment}
Generalist AI models could revolutionize the way psychological assessments are conducted. By analyzing a plethora of patient data, from spoken words to non-verbal cues, AI can assist clinicians in painting a holistic picture of a patient's psychological state. Furthermore, through continuous learning, these AI models can identify patterns and nuances often missed by the human eye, ensuring more comprehensive assessments.

\paragraph{Risk prediction}
Predictive analytics, powered by AI, can be instrumental in forecasting potential psychological risks. By parsing through historical data, genetic information, and current behavioral patterns, AI can predict susceptibility to certain disorders or the likelihood of symptom exacerbation, enabling timely interventions.

\paragraph{Diagnosis}
Speed is of the essence in many psychological crises. AI models, trained on vast datasets, can assist in rapidly analyzing patient information to provide preliminary diagnoses. This rapid assessment can then guide professionals in crafting immediate intervention strategies or treatments. Also the AI model, trained using large-scale data, is expected to accurately identify medical conditions, aiming to reduce the occurrences of misdiagnosis and missed diagnosis.

\paragraph{Treatment (cognitive behavioral therapy, communication dialogue, companionship)}
\begin{itemize}
    \item Cognitive Behavioral Therapy (CBT): AI-driven platforms can offer personalized CBT sessions, adjusting strategies based on real-time feedback and progression.
    \item Communication and Dialogue: Virtual AI therapists can facilitate round-the-clock communication, providing support during moments of acute distress.
    \item Companionship: AI companions can offer a sense of comfort, especially to individuals battling isolation, helping them navigate emotional terrains.
\end{itemize}

\paragraph{Behavior recognition}
Harnessing AI's pattern recognition capabilities can be transformative in identifying critical behaviors. Whether it's detecting early signs of impulsiveness in someone prone to self-harm or identifying potential choking hazards for patients with certain disorders, AI can provide real-time alerts and interventions.

\paragraph{Supervision and continuous care}
Post-treatment supervision is paramount in ensuring long-term well-being. AI-powered systems can monitor patients' progress, adherence to medication or therapy regimes, and any potential relapses. By providing real-time insights to caregivers or medical professionals, these systems ensure that individuals receive timely support and care adjustments.

\subsection{Generalist AI across varied psychological scenarios}
The integration of generalist AI into the realm of psychology extends far beyond traditional therapeutic spaces. The capability of AI to adapt and function across varied environments underscores its versatility. This chapter delves into the prospective roles of AI in diverse settings and the potential benefits it could usher into the field of psychology.

\paragraph{Hospital}
In a hospital setting, the precision and efficiency of AI can be monumental.
\begin{itemize}
    \item Patient Intake: AI can streamline the intake process, quickly assessing patients' psychological states, prioritizing cases based on urgency, and providing preliminary insights to medical professionals.
    \item Treatment Recommendations: By analyzing patients' medical history and current psychological assessments, AI can suggest potential treatment pathways, enhancing the decision-making process for clinicians.
    \item Post-treatment Monitoring: AI tools can assist in tracking patients' recovery, noting any anomalies or potential relapse indicators.
\end{itemize}

\paragraph{Community}
AI's role in community settings is multifaceted, encompassing both prevention and intervention.
\begin{itemize}
    \item Mental Health Awareness Campaigns: AI can analyze community-specific data to design targeted mental health awareness campaigns, addressing unique regional challenges.
    \item Early Detection: Through community health drives and surveys powered by AI, early signs of psychological distress can be identified, ensuring timely interventions.
    \item Support Groups: AI can help curate and manage community-based support groups, connecting individuals with similar experiences and facilitating therapeutic discussions.
\end{itemize}

\paragraph{Family}
Within the family setting, AI can act as a bridge, fostering communication, understanding, and support.
\begin{itemize}
    \item Home-based Therapy: AI-driven platforms can offer personalized therapy sessions, making psychological support accessible from the comfort of one's home.
    \item Conflict Resolution: AI tools can assist families in navigating conflicts, providing communication strategies and promoting understanding.
    \item Caregiver Support: For families with members undergoing psychological treatments, AI can offer resources, coping mechanisms, and real-time assistance to caregivers.
\end{itemize}

\paragraph{School}
Schools, as primary developmental environments, can greatly benefit from AI's capabilities.
\begin{itemize}
    \item Student Monitoring: AI systems can analyze students' behaviors, academic performances, and interactions to detect early signs of psychological distress.
    \item Personalized Counselling: AI-driven platforms can offer preliminary counseling sessions, guiding students through academic stresses, peer issues, or personal challenges.
    \item Educational Programs: AI can curate psychological education modules, teaching students about mental health, self-care, and resilience.
\end{itemize}

\paragraph{Social media}
In the digital age, the impact of social media on psychology is undeniable, making its intersection with AI especially pertinent.
\begin{itemize}
    \item Mental Health Check-ins: AI tools can analyze users' posts, comments, and interactions to detect signs of distress or potential cries for help, subsequently offering resources or alerting loved ones.
    \item Positive Content Promotion: AI algorithms can promote positive, uplifting content, fostering a healthier online environment.
    \item Bullying and Harassment Detection: AI can identify and mitigate instances of online harassment, ensuring safer digital spaces for users.
\end{itemize}

\subsection{Tailoring psychological Generalist AI across age groups}
\paragraph{Children: Early Detection of Psychological Issues}
Childhood is a foundational period, and timely psychological interventions can shape healthier adult lives.
\begin{itemize}
    \item Behavioral analytics: AI systems can analyze patterns in children's play, interactions, and academic performances, flagging potential developmental or psychological concerns.
    \item Interactive learning platforms: AI-driven educational tools can be designed to both teach and gauge psychological well-being, ensuring that learning and mental health go hand in hand.
    \item Parental guidance: AI can provide resources, insights, and real-time guidance to parents, helping them navigate their child's psychological needs.
\end{itemize}

\paragraph{Adolescents}
Adolescence is marked by rapid change, making tailored psychological support crucial.
\begin{itemize}
    \item Emotional well-being apps: AI-driven applications can offer adolescents a platform to understand and navigate their emotions, providing coping mechanisms for stresses tied to academics, peer pressure, identity crises, and more.
    \item Online therapy platforms: Recognizing the digital affinity of this age group, AI can power online therapy platforms, offering adolescents an accessible and less intimidating avenue for seeking help.
    \item Peer interaction analysis: AI tools monitoring online interactions can detect signs of bullying, depression, or other psychological challenges, paving the way for timely interventions.
\end{itemize}

\paragraph{Adults}
Adult life, with its myriad responsibilities, brings unique psychological challenges that AI can address.
\begin{itemize}
    \item Work-life balance tools: AI-driven platforms can assist adults in managing stress, ensuring a healthier work-life balance.
    \item Relationship and family counseling: AI systems can offer guidance on relationship challenges, parenting, and family dynamics.
    \item Mental health monitoring: Through wearable tech and mobile apps, AI can continually assess an adult's mental well-being, offering resources or interventions when necessary.
\end{itemize}

\paragraph{Elderly}
\begin{itemize}
    \item Memory and cognitive support: AI tools can offer cognitive exercises and memory aids, helping combat age-related cognitive decline.
    \item Emotional companionship: AI-powered companion robots or virtual assistants can offer conversation, alleviating feelings of loneliness and isolation.
    \item Health monitoring: Beyond mental well-being, AI tools can holistically monitor an elderly individual's health, detecting any psychological distress signs in tandem with physical health anomalies.
\end{itemize}

\subsection{Generalist AI in stages of psychological disease}
The continuum of psychological disease is complex, often marked by subtle beginnings, acute crises, and prolonged recovery phases. The application of generalist AI holds great promise in comprehensively addressing these stages, from prevention to monitoring relapses. This chapter highlights the potential interventions AI can facilitate across various disease stages, ensuring timely, effective, and empathetic care.

\paragraph{Disease prevention}
Before the onset of a psychological disorder, AI can play a pivotal role in recognizing predispositions and initiating preventive measures.
\begin{itemize}
    \item Lifestyle recommendations: Based on individual data and wider psychological research, AI can suggest lifestyle choices that optimize mental well-being.
    \item Regular mental health screenings: AI-powered platforms can offer regular mood and mental health assessments, ensuring early detection of distress.
    \item Educational content: AI systems can curate and recommend informative content on mental well-being, bolstering public awareness and preventive knowledge.
\end{itemize}

\paragraph{Early warning}
Detecting signs at the nascent stages of psychological distress can drastically improve treatment outcomes.
\begin{itemize}
    \item Behavioral pattern analysis: By analyzing behavioral data from wearables, mobile apps, or online interactions, AI can identify subtle changes indicative of emerging psychological issues.
    \item Personalized alerts: AI-driven platforms can notify individuals and their caregivers about detected changes, suggesting further assessments or interventions.
    \item Resource recommendations: Based on detected early warning signs, AI can recommend relevant self-help resources or therapeutic interventions.
\end{itemize}

\paragraph{Crisis intervention: suicide prevention}
In moments of acute distress, timely interventions are life-saving. AI holds the potential to act swiftly in such crises.
\begin{itemize}
    \item Sentiment analysis: AI tools can scan social media, messages, or other digital platforms for signs of severe distress or suicidal ideation, triggering immediate alerts to caregivers or emergency services.
    \item Immediate support chatbots: AI-driven chatbots can offer immediate emotional support, providing a bridge until human intervention becomes available.
    \item Safety planning: For individuals identified at risk, AI can assist in crafting personalized safety plans, ensuring resources and support are readily available..
\end{itemize}

\paragraph{Long-term rehabilitation monitoring}
Post-crisis, the recovery phase is crucial, necessitating consistent monitoring and support.
\begin{itemize}
    \item Daily mental health check-ins: AI-powered platforms can facilitate daily emotional and mental assessments, tracking recovery progress.
    \item Therapeutic activity suggestions: Based on recovery progress, AI can recommend therapeutic activities, exercises, or relaxation techniques.
    \item Caregiver and therapist communication: AI systems can relay progress reports to therapists or caregivers, ensuring all stakeholders are informed and aligned.
\end{itemize}

\paragraph{Disease relapse}
Ensuring that an individual doesn't revert to a previously overcome psychological challenge is vital.

\begin{itemize}
    \item Behavioral trend analysis: AI can detect patterns similar to those exhibited during prior episodes of distress, signaling potential relapses.
    \item Immediate therapeutic interventions: At signs of relapse, AI-driven platforms can offer immediate therapeutic resources or establish contact with therapists.
    \item Relapse education: AI tools can provide information on recognizing and managing relapse symptoms, empowering individuals in their ongoing recovery journey.
\end{itemize}

\section{Perspectives in practical application} \label{sec:keypoints}

As the realm of generalist AI continues to burgeon, it becomes vital to consider its practicality and implications in real-world settings. The following sections delve into the feasibility of employing large-scale AI models, the necessity of multi-domain verification, ethical concerns, and privacy considerations.

\subsection{Leveraging open source pre-trained models}
Training expansive AI models from the ground up is both resource-intensive and economically challenging, especially for institutions like hospitals or smaller enterprises. Given the substantial investments in terms of time, computing power, and finances, it's often untenable for such organizations to embark on this venture independently. The solution? Tuning based on open-source pre-trained models. This approach not only optimizes resources but also taps into the collective knowledge of the broader AI community. Given the trajectory of technological advancements, fine-tuning pre-existing models emerges as a pragmatic direction for the widespread adoption of AI capabilities.
% \textcolor{red}{(2) Retrieval-based Language Models. The language model encounters certain limitations when accessing knowledge, as its information is acquired during the pre-training phase and may not always be the most up-to-date or comprehensive. Simultaneously, training the model from scratch is a labor-intensive process. This method involves dynamically retrieving information from external knowledge bases or the internet, thereby enhancing the language model's capacity to access and manipulate knowledge. Additionally, this approach can address issues related to inaccurate content generation and hallucination problems, ultimately improving the model's ability to discern and produce reliable information.}

\subsection{The imperative of multi-domain evaluation}
The evolution and success of LLMs, evident in today's technological landscape, shine a spotlight on the potential of generalist AI models. However, potential alone isn't enough. The efficacy of these models must be tried and tested across a diverse array of domains and scenarios. By doing so, we demystify their boundaries, ascertain their capabilities, and determine their real-world relevance. This extensive verification becomes an invaluable roadmap, guiding researchers and developers alike in harnessing the true power of generalist AI. Simply put, to realize the dream of a universally applicable AI, there is no bypassing the rigorous route of multi-domain and multi-scenario verification.

\subsection{Navigating the ethical maze}
While the technical advancements in AI are noteworthy, they are not devoid of ethical quandaries. Issues of fairness, transparency, and bias persistently intertwine with AI's decision-making processes. Beyond the binary of algorithms, these concerns touch upon societal fabric, underscoring the importance of cultural and regional nuances. To ensure AI models serve everyone equitably, there's an inescapable need for continuous monitoring, adjustments, and evaluations. By remaining vigilant against inherent biases and championing transparent methodologies, the AI community can lay the foundation for ethically robust systems~\cite{joyce2023explainable}.

\subsection{Safeguarding permissions and privacy}
In an era where data is likened to gold, the stakes for protecting it are equally high. The integration of AI into various sectors necessitates stringent privacy protocols. The sanctity of personal information, be it medical records in hospitals or private chats on social media, must be preserved. Navigating the balance between AI's insatiable thirst for data and an individual's right to privacy remains one of the pivotal challenges in its practical application. As AI continues to intertwine with daily life, it becomes paramount to construct systems that prioritize user privacy, ensuring data utilization with consent and purpose.

The journey towards the seamless integration of generalist AI in practical scenarios is rife with challenges, both technical and ethical. However, by addressing these concerns head-on and leveraging the collective knowledge of the community, we move closer to an era where AI isn't just a tool but a trusted companion in various spheres of life.

\section{Conclusion} \label{sec:conclusion}
In our paper, we conduct a comprehensive review of model evaluation and the specific applications of foundation models, particularly LLMs, in the field of psychology. Our analysis emphasizes the significance of multi-faceted verification, categorizing and summarizing relevant literature by its type and application direction. Given the extensive discussions LLMs have sparked in related fields, we also collate and present various viewpoints from pertinent literature. Looking ahead, we identify potential directions for the advancement of generalist AI in psychology, focusing on four key aspects: the psychotherapy process, application scenarios, age groups, and the disease process.
Furthermore, we delve into the critical concerns regarding the practical application of LLMs. These concerns start with security and extend to advocating for open-source and transparent models, multi-domain evaluation, adherence to ethical frameworks, and addressing privacy issues to establish clear system boundaries. Our paper not only consolidates current literature but also offers in-depth analysis and exploration of future developments from multiple perspectives. This comprehensive approach aims to chart potential pathways for progress in this field and towards the development of a psychological generalist AI.

\section{Abbreviation}
\begin{itemize}
    \item AI: Artificial Intelligence
    \item APA: American Psychiatric Association
    \item CBT: Cognitive Behavioral Therapy
    \item CLEVA: Chinese Language Models EVAluation Platform
    \item GMAI: General Medical Artificial Intelligence
    \item GPT: Generative Pretrained Transformer
    \item GPAI: Generalist Psychology Artificial Intelligence
    \item LEAS: Level of Emotional Awareness Scale
    \item LLM: Large language model
    \item NLP: natural language processing
    \item ToM: Theory of Mind
    \item WMGPT: Well-Mind ChatGPT
\end{itemize}

\section{Acknowledgments}\label{sec:acknowledgments}
This work was supported by grants from the National Natural Science Foundation of China (grant numbers:72174152, 72304212 and 82071546), Fundamental Research Funds for the Central Universities (grant numbers: 2042022kf1218; 2042022kf1037), the Young Top-notch Talent Cultivation Program of Hubei Province.
Guanghui Fu is supported by a Chinese Government Scholarship provided by the China Scholarship Council (CSC).

\section{Authors Contributions}
% Conceptualization was performed by Guanghui Fu. Tianyu He, Guanghui Fu and Fan Wang conducted material preparation and literature collection. The initial draft of the manuscript was composed by XXX, XXX were responsible for review and editing... . All authors provided feedback on earlier versions of the manuscript. All authors read and approved the final manuscript.
Guanghui Fu, Bing Xiang Yang, and Jianqiang Li were instrumental in conceptualizing the idea and structure of the paper. Tianyu He, Fan Wang, Guanghui Fu, Yijing Yu, Changwei Song and Hongzhi Qi undertook the task of material preparation and literature collection. The initial draft of the manuscript was composed by Tianyu He and Guanghui Fu. Bing Xiang Yang, Jianqiang Li, Qing Zhao, Dan Luo, and Huijing Zou contributed to the article through their review and editing efforts. All authors engaged in providing feedback on previous versions of the manuscript. The final draft was read and approved by all contributing authors.

\bibliography{refs} 
\bibliographystyle{spiebib} 

\end{document}